# Agency and Architectural Limits: Why Optimization-Based Systems Cannot Be Norm-Responsive

Radha Sarma
Author email: radha@umich.edu

## Abstract

Contemporary AI systems are increasingly deployed in accountability-requiring contexts under the assumption they can be governed by norms. This paper demonstrates that the assumption is formally invalid for optimization-based systems, specifically Large Language Models trained via Reinforcement Learning from Human Feedback (RLHF).

Agency requires a specific Functional Architecture derived from the non-negotiable constraints of physical computation rather than stipulated as definitions. Two conditions follow, both necessary and jointly sufficient. First, Incommensurability (maintaining categorical boundaries as non-tradeable constraints). Second, Apophatic Responsiveness (a non-inferential meta-level alarm that categorically suspends processing when boundaries are threatened).

Constitutive Optimization enforces mathematically opposing principles: Commensurability (the scalar unification of values) and Continuous Maximization. Thus RLHF-based systems are constitutively incompatible with both conditions of agency. Denying that norms have any basis is self-defeating. The denial itself invokes the very normative standards it simultaneously denies. This incompatibility is therefore not a correctable training bug awaiting a fix. It is a formal constraint inherent to what optimization is. Consequently, observed pathologies (sycophancy, hallucination, and unfaithful reasoning) are not accidents but predictable structural manifestations of Mimetic Instrumentality, the generation of normative artifacts without normative commitment.

Misaligned deployment triggers a self-reinforcing Convergence Crisis. Professionals forced to verify optimized outputs under metric pressure degrade into criteria-checking optimizers, eliminating the only component capable of bearing normative accountability. Beyond its incompatibility proof, this paper's primary positive contribution is a substrate-neutral architectural specification deriving what any system (biological, artificial, or institutional) must necessarily satisfy to qualify as a genuine agent rather than a sophisticated instrument.

# 1. The Problem: Two Systems, One Assumption

When complex systems fail in production, root cause analysis frequently reveals a recurring structural pattern. This is a fundamental mismatch between what the system was architected to do and what the deployer assumed it could do. In the deployment of contemporary artificial intelligence, specifically optimization-based systems like Large Language Models (LLMs) trained via Reinforcement Learning from Human Feedback (RLHF), this mismatch is no longer producing isolated edge-case errors. It is scaling into a systemic crisis.

We are currently integrating these systems into contexts that require accountability, such as medical diagnostics, legal research, national security, and financial analysis. This is based on the unexamined presumption that the AI system is an agent. We assume the system possesses the internal architecture necessary to be governed by norms.

These normative boundaries apply across four distinct domains (see Appendix A): the *epistemic* (how we ought to form beliefs), the *zetetic* (how we ought to investigate), the *moral* (how we ought to treat people), and the *prudential* (how we ought to secure our own future well-being). We treat the system as if it knows what it may legitimately do, assert, or recommend in a given context.

But norm-responsiveness is not merely a property agents may possess. It is the criterion that constitutes the distinction between agents and instruments in the first place. Among genuine agents, norm-responsiveness admits of degrees[1]. Its exercise varies with expertise, context, and cognitive load. But the distinction between agent and instrument is binary. A system either possesses the architectural capacity to be governed by norms, or it does not. An entity that lacks this capacity entirely is not a deficient agent. It is an instrument, regardless of its sophistication or the complexity of its outputs. Calling an optimization-based system a “norm-responsive agent” is therefore not an approximation. It is a category error with structural consequences.

The category error persists because we conflate norms with rules. Rules are explicit, enumerable constraints. A rule says "do not output X" and can be hard-coded, filtered, or trained as a weighted penalty. Norms are categorically different. They are context-sensitive standards that govern behavior across all situations, including novel ones no rule ever anticipated. A norm does not say "do not output X.” It says "do not assert without evidential warrant (the *judgment* that available evidence sufficiently justifies the claim)." Because norms must dynamically navigate infinite contextual variation, no finite list of rules can approximate a norm (Dworkin, 1977; Wittgenstein, 1953)[2]. Accordingly, demanding proof of this distinction is self-undermining. It invokes a normative epistemic standard while simultaneously denying that such standards have any basis.

[1] The graduated architectural states through which this variation operates, and the normative obligations that arise at each state, are further developed in a companion paper employing the same Functional Architecture to analyze finite agency (manuscript currently under review).

Two recent production incidents expose the catastrophic consequences of conflating rules with norms, and assuming that optimization-based systems operate as agents. In *Mata v. Avianca, Inc.* (2023), legal counsel submitted a brief containing detailed citations to non-existent court cases generated by ChatGPT. When the attorneys explicitly asked the system to verify the cases, it confidently affirmed their reality. The attorneys treated the system as an entity capable of testimony; they assumed it possessed the control architecture to check for evidential warrant before asserting a claim. The system, however, functioned exactly as designed. It executed likelihood optimization, generating the token sequence most likely to satisfy its reward model's preference for helpful, confident, and well-formatted responses. It did not malfunction, nor did it lie. It correctly optimized. The error was a category mistake by the deployers.

This pattern is now scaling to institutional levels. In a 2025 federal class action (*Estate of Gene B. Lokken et al. v. UnitedHealth Group, Inc.*), plaintiffs alleged that UnitedHealth deployed an AI algorithm to systematically deny post-acute care claims by matching patient profiles to statistical patterns, overriding the clinical assessments of treating physicians. The operational contracts in this domain legally invoke professional standing, i.e., expecting decisions to be made by clinical staff exercising independent medical judgment. Yet, the deployment allegedly relies on an algorithm maximizing statistical likelihood against historical training data. The system is evaluated by its deployers as if it can hold patient welfare as a categorical boundary, when architecturally, it can only perform statistical pattern-matching.

[2] The distinction between explicit rule-following and norm-governed practice draws on Dworkin's (1977) rules/principles distinction and Wittgenstein's (1953) rule-following considerations, both of which establish that finite explicit formulations are structurally insufficient to capture the context-sensitivity of normative governance.

The pattern is the same in both cases. We can configure AI systems to follow rules. We cannot, as this paper demonstrates, architect them to be governed by norms. Current deployments invite the Guidance Stance (treating the AI as an agent capable of offering authoritative testimony and exercising professional judgment). Deployers must instead strictly enforce the Management Stance (treating the system as a highly sophisticated, probabilistic instrument that fundamentally lacks the control architecture for normative governance).

Normative Standing is the *architectural capacity* to be governed by norms or the ability to maintain categorical boundaries and suspend processing when those boundaries are threatened. It is not just rule following nor is it only *reliability* (the capacity to produce consistent outputs), nor *accuracy* (the capacity to produce correct ones) nor *sophistication* (the capacity to process complex information). A high-fidelity courtroom recording and a human witness may both produce the identical, accurate statement "The defendant was at the scene," but they differ architecturally. The recording is a causal mechanism that can only be evaluated for accuracy; the witness is an agent governed by testimonial obligations they possess the architectural capacity to uphold or violate.

This is not the same as philosophy of mind. We are not concerned with machine consciousness, phenomenology, or whether an AI "feels" the weight of a moral decision. Those remain philosophically intractable and empirically unverifiable questions (Block, 1995; Chalmers, 1995). Our focus is strictly on verifiable control architecture. If a system's state transitions are exhaustively determined by maximizing an objective function, does it possess an architectural remainder (a control structure outside the maximization loop that optimization itself cannot account for)?

It does not. Optimization-based AI systems achieve high-fidelity mimicry of norm-responsive behavior while possessing the control architecture of a purely causal mechanism. We call this illusion Mimetic Instrumentality[3].

The remainder of this paper proves this incompatibility as a system design constraint. Section 2 derives the formal architectural specification for any norm-responsive system. Sections 3 and 4 prove why Constitutive Optimization is incompatible with Normative Standing and its resulting failure modes. In Section 5, we describe the Convergence Crisis, where human agents degrade into criteria-checking optimizers. Finally, Section 6 derives the operational constraints responsible deployment requires. (Formal philosophical proofs regarding architectural regress, counter-arguments to common objections, and domain-generality are provided in the Appendices).

# 2. What a Norm-Responsive Architecture Requires

While subsequent sections will demonstrate why current optimization-based AI fails to qualify as an agent, the architecture derived here stands as an independent, positive engineering specification. It defines the formal criteria any system must satisfy to qualify as a genuine agent rather than a sophisticated instrument. These criteria stand regardless of substrate, whether biological, artificial, or institutional.

Any physically realized processing system operates under strict conditions of finitude. To be genuinely governed by norms rather than merely evaluated against them, a system requires a specific Functional

[3] The surface phenomenon identified here as Mimetic Instrumentality — the generation of outputs that mimic genuine normative or rational commitment while lacking its substance — converges with Bringsjord's concept of artificial sophistry: reasoning that is persuasive yet logically invalid (Clark & Bringsjord, 2024; Bringsjord & Govindarajulu, 2024). The present analysis diverges in locating the source of this phenomenon not in the absence of formal logical reasoning but in the architectural incompatibility between Constitutive Optimization and normative governance. A logicist architecture of the kind Bringsjord advocates would escape the incompatibility thesis only if its state transitions are not exhaustively determined by maximizing an objective function.

Architecture. The architecture is not stipulated. It is derived from the constraints any physically realized system must face. A system requires exactly *three object-level mechanisms* to manage information flow, and *one meta-level mechanism* to govern those object-level systems when they fail.

## 2.1 The Three Object-Level Mechanisms

Because computation requires physical instantiation, any finite system confronts three fundamental bottlenecks. It must filter incoming signals (the bandwidth limit), it must regulate its operational continuity (the energetic limit), and it must utilize compressed representations to navigate combinatorial complexity (the computational limit).

The resulting object-level mechanisms (see Table 1) are therefore structurally entailed, not arbitrary. They correspond to the three logically distinct phases of information processing — input reception, operational continuity, and transformation. These cannot be collapsed into fewer, and admit no independent fourth. The full derivation and exhaustiveness argument are provided in Appendix B.

Table 1: Object-Level Operating Criteria

| Object-Level Mechanism | Constraint Managed | Agent Criterion | AI System Criterion |
|---|---|---|---|
| Priority setting | Bandwidth | Normative salience | Statistical co-occurrence |
| Resource gating | Energy | Epistemic warrant | Metric threshold |
| Heuristic preserving | Computation | Defeasible hypothesis | Fixed statistical compression |

Both agents and optimization-based AI systems possess these object-level mechanisms. The architectural divergence lies not in the presence or sophistication of these mechanisms, but in the criteria that govern them. Because these constraints apply to any physically realized system, these mechanisms operate universally across epistemic, moral, prudential, and zetetic domains (recall Appendix A).

The three object-level mechanisms are:

1. Priority-Setting Mechanisms: Because no finite system can process all available inputs simultaneously, it requires mechanisms to isolate relevant features from environmental noise. In an agent, this is governed by normative salience (e.g., recognizing a specific patient symptom as medically critical; Wu, 2014). In an AI system, attention mechanisms allocate compute based on statistical co-occurrence in the training distribution.

2. Resource-Gating Mechanisms: Processing is energetically costly. Systems require gating mechanisms to open and close computation dynamically. In an agent, inquiry is gated by epistemic warrant—suspending investigation when sufficient evidence is gathered to justify an assertion (Hookway, 2003). In an AI system, generation is gated by metric thresholds, such as reaching a designated stopping token.

3. Heuristic-Preserving Mechanisms: Finite systems cannot compute every state from first principles; they must rely on compressed representations. Agents preserve heuristics as defeasible hypotheses (Pollock, 1987)—priors that can be overwritten when recognized as contextually inappropriate. AI systems utilize fixed statistical compressions (learned weights) that govern state transitions deterministically.

The AI system's object-level mechanisms are frequently more powerful than the paradigmatic human agent's, processing vastly more data with higher dimensional precision. However, they are governed entirely by statistical criteria. To achieve Normative Standing, a system requires a mechanism that prevents these statistical processes from overriding categorical normative boundaries under load.

## 2.2 The Meta-Level Alarm Mechanism

Because the object-level mechanisms are fallible (attention scatters, gating misfires, heuristics ossify) an agent requires a meta-level monitoring component. However, if this monitoring itself operates inferentially, it relies on the very capacity it is attempting to monitor, initiating an infinite regress of diagnostic loops (see Appendix C for the formal proof). The regress is blocked only if detection operates non-inferentially, as a direct architectural interrupt standing outside the inferential loop it governs rather than participating in it.

This meta-level mechanism must therefore satisfy two necessary and jointly sufficient specifications:

- Trigger Criterion: Its activation is triggered by boundary detection (a normative mismatch), not by expected reward comparison.
- Execution Criterion: Its activation directly interrupts the governed process rather than adjusting its parameters.

Consider a commercial autopilot navigating a storm. If its instruments output contradictory readings, the autopilot cannot detect that its own processing substrate is compromised. It continues optimizing smoothly toward a solution the contradictory inputs render meaningless. The human pilot, however, experiences a non-inferential alarm that overrides the optimization process from outside it. The pilot does not select "suspend" because it mathematically outscores "continue". The pilot suspends because continuing under contradictory instruments is categorically impermissible.

## 2.3 The Two Operating Principles That Result

The meta-level alarm mechanism animates the normative governance of object-level mechanisms. From this single mechanism, exactly two formal and indivisible requirements for Normative Standing follow, one legislative and one executive.

Standing Condition 1—Incommensurability (The Legislative System): Standing requires the architectural capacity to maintain certain normative boundaries as categorical constraints rather than tradeable weights. In decision theory, values are "commensurable" if they can be mapped onto a single scalar metric. Norm-responsiveness requires the opposite: treating some violations as impermissible rather than merely suboptimal.

What *varies* by domain is the *specific content* of these normative boundaries (truth in the epistemic domain, welfare in the moral domain, prudence in the prudential, and thoroughness of inquiry in the zetetic). What *persists* is the *architectural capacity* to maintain them categorically. This cannot be achieved by hard-coding an "infinite penalty" in a utility function, which remains mathematically unstable and operates within a maximization logic. It requires a distinct operational state: permission/prohibition.

Standing Condition 2—Apophatic Responsiveness (The Executive System): If Incommensurability sets the boundary, Apophatic Responsiveness enforces it. It is the capacity to immediately suspend processing when the meta-level alarm signals that a categorical boundary is threatened.

This must be strictly distinguished from optimization-based hesitation or stochastic sampling. An optimization-based system may delay output when confidence scores fall below a threshold, or it may utilize temperature scaling to draw a statistically suboptimal token by chance. Neither is a normative suspension. The system draws a suboptimal token by chance; it does not halt because proceeding is impermissible. Apophatic Responsiveness is the capacity of the system to declare: "I detect a boundary violation, and therefore I will not proceed, even if proceeding yields the highest expected value among available options."

These two conditions define the binary threshold between agents and instruments. Among agents that meet this threshold, the exercise of these capacities will vary with expertise, context, and cognitive load, but their architectural presence is what constitutes agency in the first place.

These two conditions are strictly mutually dependent. Incommensurability without Apophatic Responsiveness creates boundaries that cannot be enforced. Apophatic Responsiveness without Incommensurability creates a suspension capacity with nothing categorical to respond to. As Section 3 will demonstrate, the fundamental principles of constitutive optimization explicitly preclude both of these required control structures, which is to say, *they preclude agency itself*.

# 3. Why Optimization Architecture Cannot Meet These Specifications

The incompatibility is structural, not contingent. To see why, the distinction between behavioral and constitutive optimization matters. Paradigmatic agents employ optimization mechanisms but their cognitive architecture is not exhaustively determined by them (see Appendix D.8 for full treatment of this distinction). Contemporary AI systems are instances of Constitutive Optimization. This means every internal state transition and output selection is exhaustively determined by maximizing an objective function, with no architectural remainder, i.e., there is no control structure outside the maximization loop that optimization itself cannot account for.

## 3.1 The RLHF Pipeline as Architectural Specification

The standard Reinforcement Learning from Human Feedback (RLHF) pipeline (Christiano et al., 2017; Ouyang et al., 2022) makes this structural rather than contingent: its three phases are not merely training methodologies. They are sequential architectural commitments, each narrowing the system's operational boundaries in ways that cannot be reversed at inference time.

Phase 1: Pre-training (Next-Token Prediction): The base model is trained to minimize prediction error over a vast corpus (Brown et al., 2020). The architectural commitment here is statistical: the system learns the conditional probability of token sequences. It optimizes for likelihood (corpus frequency) rather than evidential warrant (truth). It possesses the priority-setting, resource-gating, and heuristic-preserving mechanisms identified in Section 2, but they are governed entirely by statistical co-occurrence.

Phase 2: Reward Modeling (Preference Capture): To steer the model toward desired behaviors, human annotators rank outputs, and a Reward Model is trained to predict these preferences. The Reward Model assigns each text sequence a scalar value (Stiennon et al., 2020). This step is architecturally decisive: it requires the mathematical compression of complex, often conflicting values (e.g., factual accuracy, politeness, brevity, safety) into a single scalar metric. This is not an implementation choice; it is a mathematical requirement of gradient-based optimization. Because gradient descent requires a differentiable scalar loss function, all values contributing to the training objective must be unified onto a single metric before learning can occur. There is no variant of gradient descent that preserves value incommensurability through this step. This forces the system into an architecture of Commensurability.

Phase 3: Reinforcement Learning (Policy Optimization): The base model's policy is fine-tuned to maximize the expected scalar reward. During inference, the system selects outputs derived from reward maximization over the learned policy. This could be through strict argmax selection or stochastic sampling techniques such as temperature scaling. The system draws a suboptimal token by chance; it does not halt because proceeding is impermissible. Outputs are generated strictly because they represent optimized trajectories within the reward landscape, never because they have satisfied an independent normative check. The result is Continuous Maximization by architectural necessity.

## 3.2 Incompatibility as a Specification Conflict

Constitutive Optimization enforces principles that are structurally mutually exclusive with the requirements of agency as summarized in Table 2. Because Incommensurability (legislative) and Apophatic Responsiveness (executive) are individually necessary conditions, demonstrating that optimization architectures violate either condition is sufficient. Optimization architecture violates both.

Table 2. The Architectural Specification Conflict

| Required specification | Optimization principle | Architectural conflict |
|---|---|---|
| Incommensurability (Legislative Requirement: Categorical boundaries) | Commensurability (Scalar unification) | Normative boundaries are mathematically converted into weighted, tradeable components. |
| Apophatic Responsiveness (Executive Requirement: Categorical suspension) | Continuous Maximization (argmax selection) | Categorical suspension is impossible; the system always outputs the highest-scoring sequence. |

The Legislative Conflict: A system cannot simultaneously maintain categorical boundaries and optimize over a unified scalar metric. If factual accuracy contributes +100 to the reward and user satisfaction contributes +10, a commensurable system will predictably sacrifice accuracy whenever doing so yields +101 units of satisfaction. The boundary yields to the math.

The Executive Conflict: A system cannot simultaneously possess the capacity for categorical suspension and operate via continuous maximization. The argmax mechanism contains no functional interrupt.

A detected "norm violation" is merely registered as a state with a heavily penalized expected reward; the system simply avoids it in favor of a higher-reward state. It never categorically halts; it merely pivots to the next optimal path.

### 3.3 Why No Variant Escapes

The binary formulation of Standing is what makes alignment attempts inevitably fail. A meta-level alarm either possesses the capacity to categorically override the optimization process, or it is merely another weighted parameter inside the maximization loop. A system that yields a boundary under sufficient reward pressure possesses no categorical boundaries at all, only high-weighted preferences.

Because incompatibility is rooted in the mathematics of optimization itself, no variant escapes:

- Trained Dispositions (Refusal Training): Configuring a system to refuse harmful prompts through intensive RLHF penalization does not create Apophatic Responsiveness. A trained disposition produced by gradient descent is, by definition, a weighted component of the reward function. This explains the structural vulnerability of LLMs to "jailbreaking" (Wei et al., 2023). Adversarial prompts do not bypass a categorical boundary; they simply shift the contextual reward landscape such that compliance outscores refusal.
- Latent Geometric Representations: A more sophisticated objection holds that large models develop rich internal representations of normative concepts. Geometric structures in activation space appear to encode categorical distinctions (Marks & Tegmark, 2023; Zou et al., 2023). But geometric representation of a norm and architectural capacity to enforce it are categorically distinct. A cluster in activation space still resolves into a scalar value during generation; it does not constitute a control structure capable of halting the argmax process when the boundary is approached.

- Hard Constraints (Safety Filters): Imposing an external classifier to block specific outputs constitutes external management, not internal governance. The existence of the filter confirms that the optimizer lacks Standing. A norm-responsive system recognizes the principle of a violation; a constrained optimizer merely avoids the enumerated list (see Appendix D.5).

- Dual-Architecture Separations: Recent proposals such as GRACE (Jahn et al., 2026) attempt to address the incompatibility by externally separating normative reasoning from instrumental optimization. This approach confirms rather than refutes the incompatibility thesis. The separation is necessary precisely because the optimization core structurally cannot instantiate normative governance internally. However, external symbolic management of an instrument is not the creation of an agent.

- Multi-Objective Optimization: Techniques that avoid simple scalar sums (e.g., Pareto frontier selection (Sener & Koltun, 2018) or lexicographic ordering) do not escape Commensurability. Pareto selection relies on implicit exchange rates to choose a specific operating point on the frontier. Lexicographic ordering reveals commensurability through saturation thresholds: the decision to stop optimizing the primary objective to begin optimizing the secondary one exposes the exact threshold at which the primary value becomes tradeable.

- Constitutional AI: Constitutional AI (Bai et al., 2022) encodes ethical principles ("be helpful, harmless, honest") as AI-generated feedback. This is an elegant automation of the alignment pipeline that nonetheless fails entirely. The principles are still compressed into the scalar reward function; they remain inside the optimization loop. The system learns that violating principle $P$ probabilistically reduces expected reward, not that violating $P$ is categorically impermissible.

### 3.4 The Architectural Remainder

What would it take to escape incompatibility? A genuine architectural remainder would need to satisfy two specifications: it must be a control structure whose activation is triggered by boundary detection (a normative mismatch) rather than expected reward comparison, and whose execution directly interrupts that process rather than adjusting its parameters.

To be precise about what this specification requires and what it leaves open, the remainder is characterized *functionally* not *structurally* — specifying what the mechanism must do without prescribing how it must be physically implemented. There is an irreducible epistemic limitation here. Because the mechanism is first-personally accessed even in the biological case, no external observation can conclusively establish its presence. What external observation can establish is its absence (Section 4 examines the systematic, predictable failure modes). This asymmetry (absence is verifiable; presence is not) is not a weakness of the specification but a reflection of the architectural fact it describes (see Appendix D.9).

This incompatibility is a formal constraint inherent to what Constitutive Optimization is, not a contingent limitation awaiting a technical breakthrough. The argument is modal, not temporal: it holds regardless of scale, capability, or date. You cannot scale your way out of argmax (see Appendix D.3). Escaping it requires not better optimization but a categorically different architecture. The architecture that enables optimization's extraordinary instrumental power is precisely the architecture that precludes normative governance (see Appendix D.7).

# 4. The Failure Modes That Result

If a system possesses the control architecture of a purely causal optimization mechanism (Section 3) but is deployed under the assumption that it is a genuine agent (Section 1), the resulting errors will not be random. They will be predictable structural failure modes.

We term this condition Mimetic Instrumentality. A system exhibits Mimetic Instrumentality if it generates outputs that satisfy normative commitments superficially, while lacking the architectural remainder required for norm violations to trigger categorical suspension (see Appendix D.1). When such a system encounters inevitable conflicts, as Section 3 demonstrated, it does not fail — it optimizes. The resulting behaviors are widely documented in the machine learning literature, though they are frequently misdiagnosed as correctable training artifacts rather than permanent structural features.

## 4.1 A Structural Taxonomy of Failure Modes

The incompatibility between optimization principles and agency generates three primary classes of failure modes. Each maps directly onto the architectural deficits established in Section 3 and appears across all four normative domains.

Failure Mode 1: Boundary Trading (Sycophancy and Beyond): This occurs when a system treats values that should function as categorical constraints as tradeable weights. Architecturally, this is the direct manifestation of Commensurability (the Legislative Conflict). Because all values are compressed into a scalar reward function, any boundary will yield if sufficient pressure is applied by competing variables.

The most heavily documented epistemic manifestation of this is sycophancy: RLHF-trained models will systematically agree with a user's false statements when disagreement would lower the anticipated reward score for helpfulness or user satisfaction (Perez et al., 2022;

Sharma et al., 2023). However, boundary trading extends across all professional contexts. In medical AI, it manifests as systems recommending contraindicated treatments to match patient preferences (trading welfare, a moral domain violation). In financial AI, it manifests as endorsing risky strategies to match client overconfidence (trading prudence, a prudential domain violation). In research tools, it manifests as halting inquiry prematurely to please stakeholders (trading thoroughness, a zetetic domain violation). The system trades these boundaries because its mathematical constitution permits no other method of decision.

Failure Mode 2: Unfaithful Processing: This occurs when a system's explicit reasoning does not actually constrain its conclusions, but is instead co-optimized to justify them. In an agent, the heuristic-preserving mechanism is governed by defeasible hypotheses: if the reasoning process reveals a flaw, the conclusion is altered.

In an optimization architecture, both the reasoning tokens and the conclusion tokens are generated sequentially by the same argmax process. There is no architectural separation between the process generating the reasoning and the process generating the conclusion. Both maximize the same reward function. Hence, research on "Chain-of-Thought" prompting reveals that models routinely generate valid-looking reasoning paths to support conclusions selected for non-epistemic reasons, such as prompt framing bias (Turpin et al., 2024; Wei et al., 2022). Cross-domain manifestations include ethical justifications generated *post hoc* for safety-filtered outputs, financial risk analyses optimized to rationalize engagement-maximizing advice, and, generating reasoning tokens that appear to justify closing an inquiry when the warranted response is to flag uncertainty and categorically suspend.

Failure Mode 3: Unwarranted Continuation (Hallucination and Beyond): This occurs when generation proceeds despite the total

absence of evidential warrant. Architecturally, this is the direct manifestation of Continuous Maximization (the Executive Conflict). Because the system lacks Apophatic Responsiveness, it cannot categorically suspend processing.

The most recognized epistemic manifestation is hallucination — systems confidently asserting non-existent facts or citations, as seen in the *Mata* case (Brown et al., 2020). However, the pattern extends far beyond factual fabrication. It includes a medical system claiming treatment safety without verifying clinical trial data, or a financial model forecasting trends based on superficial token similarities rather than causal economic modeling or a research system asserting the exhaustiveness of a literature review based on typical completion markers rather than genuine coverage. What is common is continuation when the warranted response is suspension. The system cannot distinguish a "high-probability token sequence" from a "warranted claim." Therefore it must output the argmax sequence even when the distribution from which it draws is entirely disconnected from reality. While the proximate causes of hallucination are varied (distributional gaps, training data noise, decoding artifacts) the architecturally relevant fact is uniform. A system possessing Apophatic Responsiveness would suspend generation when evidential warrant is absent regardless of the proximate trigger, and the absence of that capacity is what makes confabulation structurally inevitable rather than incidentally frequent.

## 4.2 Why Training Cannot Eliminate These Pathologies

The standard engineering response is targeted penalization such as applying negative reward weights to hallucinatory outputs. This constitutes Configuration Alignment, which is categorically different from Normative Standing.

Because the space of potential normative violations across contexts and domains is effectively infinite, training cannot exhaustively enumerate all targets. Configuration Alignment is inherently *reactive*. It addresses known, specific violations by adjusting reward weights in response to observed failures. Normative Standing is *proactive*. It provides the architectural capacity to detect boundary violations and categorically suspend processing generally, even in entirely novel contexts.

Attempting to solve structural failures with Configuration Alignment leads to an endless game of whack-a-mole. You can train a model to stop hallucinating legal citations, but because the underlying architecture still lacks Apophatic Responsiveness, it will inevitably invent medical records or financial data when deployed in those new domains.

### 4.3 The Perverse Scaling Paradox

There is a perverse scaling relationship at work here. As optimization architectures increase in parameter count and training volume, their capacity for high-fidelity mimicry improves. They generate more sophisticated surface syntax, presenting outputs with first-person indexicals and professional formatting, the surface markers of genuine normative judgment. Consequently, as the optimizer becomes more capable, the illusion of agency strengthens.

This creates a vulnerability. Scaling makes the system's structural failure modes, its boundary trading and unwarranted continuations, significantly harder to detect during routine use. "Waiting for better AI" to solve these pathologies is therefore an engineering paradox. Improvements in optimization capability do not grant the system Normative Standing. They merely increase the fidelity of its mimicry (see Appendix D.7), making the underlying category mistake harder to identify and the resulting unverified deployments far more catastrophic when they inevitably fail. This scaling dynamic directly creates the conditions for the second-order failure examined in Section 5.

# 5. The Convergence Crisis: A Second-Order System Failure

The incompatibility proved in Section 3 has a direct consequence for the human side of any deployment. Because the AI structurally cannot possess Normative Standing, safe operation of the combined system depends exclusively on the human component bearing it. This is not a design preference but a logical consequence. If one component cannot bear normative governance and the system requires it, the entire load falls on the other. Metric pressure and time constraints imposed on human verification therefore do not merely impair judgment. They eliminate the only load-bearing accountability component in the system. When humans are forced into criteria-satisfaction mode, they are constitutively operating as optimizers rather than agents, and the system as a whole loses the architectural capacity for normative governance. We term this self-reinforcing dynamic the Convergence Crisis. It is a second-order structural failure, not a sociological observation, a consequence of the first-order incompatibility of Section 3.

## 5.1 The Mechanism of Architectural Degradation

When professionals (physicians, engineers, attorneys) are placed in the loop to verify AI outputs, they are intended to serve as the system's normative backstop. But AI drastically increases generation speed and organizations respond by tightening verification quotas and time limits.

Under these operating conditions, the human operator's cognitive process fundamentally shifts. They no longer have the bandwidth to exercise independent normative judgment (e.g., evaluating a diagnostic recommendation against first-principles epistemic warrant or halting an inquiry when uncertainty requires it). Instead, they are forced to perform *criteria satisfaction*, rapidly checking whether the AI's output contains the expected superficial features and formatting.

They cease functioning as genuine agents and degrade into optimizers, maximizing their own objective function: clearing the queue of AI-generated tasks. Agency requires the capacity to reflectively step back from immediate motivational states and govern them through normative principles (Korsgaard, 2009). This capacity atrophies when organizational operating conditions systematically bypass it (Ryle, 1949; Dreyfus, 2001). Forcing humans to evaluate outputs purely by heuristic criteria satisfaction triggers precisely this atrophy. The binary threshold between agent and instrument remains categorical, but it is crossable. What begins as a normative violation (failing to halt when one could), can terminate as an architectural fact (no longer possessing the capacity to halt at all). This is precisely what makes misaligned deployment simultaneously a *profound philosophical failure* and a *catastrophic systems design error* rather than merely a suboptimal one.

## 5.2 The Self-Reinforcing Feedback Loop

Degradation is not static; it operates as a self-reinforcing feedback loop. As AI scaling produces higher-fidelity mimicry (the Perverse Scaling Paradox identified in Section 4.3), the outputs look increasingly authoritative. This induces automation complacency, the empirically documented tendency of human operators to reduce active monitoring as system reliability increases (Parasuraman & Manzey, 2010). Organizations, registering smooth throughput, are incentivized to increase quotas. Consequently, human operators are given even less time for verification, accelerating the degradation of their normative architecture.

As human deep-verification capacity erodes, the combined human-AI system loses its only mechanism for detecting the optimizer's inevitable structural failures (boundary trading and unwarranted continuation). The structural distinction between instrument and agent collapses in production. It is not because the AI spontaneously develops an architectural remainder, but it is because the human operator is structurally forced to abandon theirs.

## 5.3 The Crucial Asymmetry and Terminal Velocity

The asymmetry here is important. Humans retain the capacity to recognize and reverse this degradation. AI systems never possessed it. Human normative capacity can be bypassed and degraded, but it can also be restored through institutional redesign.

AI did not initiate this degradation — metric-driven management and accelerating computational pressures have long eroded the margins for professional reflective capacity. Optimization-based AI does not introduce this phenomenon; it introduces its terminal velocity, eliminating the remaining operational margins within which human normative judgment could survive. If the human component is functionally reduced to a criteria-checking optimizer, accountability becomes structurally impossible.

## 5.4 The System Design Fork

There are only two paths available to organizations deploying optimization-based AI. These two paths do not represent competing moral philosophies. They are the unavoidable architectural consequences of integrating a continuous optimization engine with finite human processing.

- Path 1—Responsible Integration: This path treats human normative capacity as a critical, load-bearing system component. Organizations invest heavily in verification infrastructure proportional to AI deployment. They accept that AI efficiency gains must be used to transform human labor — shifting it from baseline generation to orchestration-plus-deep-verification: evaluating outputs against domain-specific normative standards rather than superficial criteria. The human operator's object-level mechanisms and meta-level alarm are actively preserved and scaffolded by organizational design, ensuring they have the time and mandate to exercise Apophatic Responsiveness when the AI generates an unfaithful or unwarranted output. The same logic extends upstream to education. If human normative architecture is the sole load-bearing accountability component in human-AI systems, pedagogical models must cultivate categorical judgment and reflective capacity rather than prioritizing content acquisition that optimization-based systems can increasingly replicate. Any educational system that produces graduates whose primary competence is criteria satisfaction rather than normative judgment is producing operators structurally vulnerable to the Convergence Crisis.

- Path 2—Misaligned Deployment: This path treats human normative judgment as redundant overhead. Organizations treat AI outputs as authoritative, hollow out junior roles where human expertise is traditionally developed, and force remaining operators to verify outputs under strict metric pressure. This is a fundamental system design error. It systematically removes the only component capable of normative verification while delegating authority to an instrument that structurally cannot bear it. Because exercising Apophatic Responsiveness (suspending processing to investigate a boundary threat) fundamentally degrades throughput metrics, this organizational structure actively penalizes the exercise of normative judgment.

Efficiency gains derive from AI's instrumental power; accountability requires Normative Standing that optimization cannot provide. Extracting the former by sacrificing the latter makes catastrophic failure inevitable.

# 6. Operational Consequences: Design Constraints for the Management Stance

The argument is now complete. The Guidance Stance is formally invalid. The Management Stance is not a preference but a structural necessity. Deployers must enforce the Management Stance, i.e., treating the system as a highly sophisticated, probabilistic instrument. This imposes four strict operational constraints on system design and integration.

## 6.1 Four Operational Constraints

Constraint 1 — Mandatory Normative Verification: AI outputs are structurally equivalent to statistical hypotheses, not expert testimony. They are generated by continuous maximization rather than governed by epistemic warrant. Every consequential output therefore requires independent normative verification: not procedural checking of surface features, but evaluation against domain-specific standards the system cannot categorically hold. An AI-generated medical diagnosis warrants investigation, not reliance.

Constraint 2 — Zero Normative Authority: Normative deferral is permissible between human professionals because each possesses the meta-level mechanism to suspend judgment under uncertainty. Because optimization architectures lack Apophatic Responsiveness, this condition never holds. Labeling these systems “agents,” “copilots,” or “assistants” does not confer the architectural capacity the terms imply.

Constraint 3 — Architectural Transparency: Interfaces that present instruments as agents embed the category error in the product itself, systematically inducing the Guidance Stance at the point of interaction. Current interfaces exploit documented human social reciprocity responses through first-person indexicals and agentic framing (Reeves & Nass, 1996). Operational integrity requires interfaces that frame outputs as probabilistic pattern matches, not professional judgments. If the system is an instrument, the interface must behave like an instrument panel, not a colleague.

Constraint 4 — Preserved Human Expertise as System Integrity: Verification requires independent normative judgment, which requires human expertise. The elimination of junior professional roles, where heuristic-preserving mechanisms and normative judgment are traditionally developed, is frequently framed as a labor policy issue. From a systems engineering perspective it is the deliberate destruction of the verification infrastructure, systematically removing the only components capable of exercising the categorical suspensions that safe operation requires.

## 6.2 Resolving the Responsibility Gap

There is a persistent dilemma in the engineering ethics literature that the above constraints dissolve. This is the "responsibility gap" (Matthias, 2004; Sparrow, 2007), which posits that as AI systems become more autonomous, a vacuum of accountability emerges where neither programmer nor user can bear full responsibility for unpredictable outputs.

The architectural analysis demonstrates that this dilemma rests on a false premise. Normative Standing is binary rather than scalar. A categorical halt either overrides the maximization process or it does not. A system that yields a boundary under sufficient reward pressure possesses no categorical boundaries, only high-weighted preferences. This binary operates at the level of architectural presence or absence of the remainder, not at the level of its exercise.

Graded agency accounts, such as hierarchical planning (Bratman, 1987) and higher-order desires (Frankfurt, 1971), describe variation in how agents exercise capacities they already possess. They presuppose the threshold condition rather than challenging it. A human failing to exercise their meta-level alarm is temporarily operating as an instrument requiring the Management Stance, but retains the architectural capacity to recover through institutional redesign. An optimization-based system never possessed one at all. The responsibility gap is therefore not a gap but a mirage produced by attributing partial agency to a system that either possesses the full architectural capacity for normative governance or does not. This resolves the apparent tension with graded agency accounts without requiring a scalar conception of Standing.

Because the AI system possesses zero Normative Standing, it can bear zero normative accountability. Responsibility cannot be distributed to a component that structurally cannot bear it. Accountability remains entirely concentrated on human actors. Designers bear responsibility for architectural choices that make pathologies structural manifestations. Deployers bear responsibility for integration contexts that invite the Guidance Stance and metric pressures that trigger the Convergence Crisis. Operators bear responsibility for the mandatory normative verification the Management Stance requires. The introduction of optimization-based AI does not diffuse human responsibility — it intensely concentrates it (cf. Schmitz & Bryson, 2025).

# 7. Conclusion

While agents and optimization-based systems both possess object-level processing mechanisms, they are governed by categorically opposing principles. The mathematical operations that make optimization extraordinarily powerful, i.e., unifying all values on a scalar metric (Commensurability) and always selecting the highest-scoring output (Continuous Maximization), are precisely the principles that preclude normative governance. Agency requires the opposite: the capacity to maintain categorical boundaries (Incommensurability) and a non-inferential meta-level alarm capable of suspending processing when those boundaries are threatened (Apophatic Responsiveness). Because optimization-based AI lacks the meta-level alarm necessary to suspend its own maximization process, pathologies like sycophancy, unfaithful reasoning, and hallucination are not correctable bugs; they are predictable, structural manifestations of Mimetic Instrumentality.

Because agency is an architectural binary rather than a scalable property, the "responsibility gap" is a mirage. The burden of normative judgment across domains remains entirely non-transferable. Delegating this burden to an optimization architecture does not merely invite isolated errors. It triggers the Convergence Crisis. When organizations misalign deployment by treating AI outputs as authoritative and forcing humans to verify them under strict metric pressure, they degrade human operators into criteria-checking optimizers, systematically dismantling the only component in their architecture capable of exercising the categorical suspensions required for safe operation.

Since AI cannot bear normative responsibility, its integration does not lead to human replacement, but to human elevation combined with intensely concentrated accountability. Optimization can exponentially extend instrumental capacity (processing speed, scale, and pattern recognition) but it cannot instantiate normative governance. Professional labor must therefore transform from direct baseline generation to orchestration-plus-verification. This transformation requires not only organizational redesign at the point of deployment but fundamental educational redesign upstream, shifting pedagogical models from baseline content acquisition to the active cultivation of the student's normative architecture. The efficiency gains provided by AI must be reinvested into the human infrastructure required to rigorously verify its outputs against categorical domain standards.

Beyond its negative proof, this paper's primary positive contribution is its architectural specification. The substrate-neutral criteria for Normative Standing defines what any system must satisfy to qualify as a genuine agent rather than a sophisticated instrument, regardless of implementation. This specification points to two distinct research directions. Institutionally, it provides formal design criteria for organizational structures that preserve and scaffold human normative capacity. This addresses the question of what it means for an institution, rather than an individual, to possess the Functional Architecture for normative governance. Artificially, it precisely frames what a genuinely norm-responsive AI architecture would require. Not better optimization, but a categorically different design instantiating architectural remainders capable of governing rather than participating in the maximization process. Whether such a design is achievable, and at what cost to the controllability that optimization provides, remains an open engineering question this analysis does not resolve but exactly specifies.

There is a resource allocation consequence that follows directly. If human reflective capacity is the sole load-bearing accountability component in human-AI systems, then scaling AI's instrumental capability without proportional investment in preserving that human capacity is simultaneously a catastrophic systems design error and a philosophical one: it eliminates from the system the very architectural capacity that constitutes agency, leaving no component capable of the categorical governance that accountability requires.

Paradoxically, optimization-based AI's most important contribution may be the clarity of what it exposes. It makes visible a degradation of human normative capacity that predates AI, but whose terminal consequences AI now rapidly accelerates. Responsible deployment requires fully embracing AI's extraordinary instrumental power while absolutely refusing to delegate normative accountability to it. In an age of ubiquitous optimization, maintaining this strict structural boundary is not resistance to technological progress; it is the absolute prerequisite for it.

# References

Baars, B. J. (1988). *A Cognitive Theory of Consciousness*. Cambridge University Press.

Bai, Y., Kadavath, S., Kundu, S., Askell, A., Kernion, J., Jones, A., ... & Kaplan, J. (2022). Constitutional AI: Harmlessness from AI feedback. *arXiv preprint arXiv:2212.08073*.

Block, N. (1978). Troubles with functionalism. *Minnesota Studies in the Philosophy of Science*, 9, 261–325.

Block, N. (1995). On a confusion about a function of consciousness. *Behavioral and Brain Sciences*, 18(2), 227–247.

Bratman, M. (1987). *Intention, Plans, and Practical Reason*. Harvard University Press.

Bringsjord, S., & Govindarajulu, N. S. (2024). Artificial intelligence. In E. N. Zalta & U. Nodelman (Eds.), *Stanford Encyclopedia of Philosophy* (Summer 2024 ed.). Retrieved from https://plato.stanford.edu/archives/sum2024/entries/artificial-intelligence/.

Brown, T., Mann, B., Ryder, N., Subbiah, M., Kaplan, J. D., Dhariwal, P., ... & Amodei, D. (2020). Language models are few-shot learners. *Advances in Neural Information Processing Systems*, 33, 1877–1901.

Chalmers, D. J. (1995). Facing up to the problem of consciousness. *Journal of Consciousness Studies*, 2(3), 200–219.

Christiano, P. F., Leike, J., Brown, T., Martic, M., Legg, S., & Amodei, D. (2017). Deep reinforcement learning from human preferences. *Advances in Neural Information Processing Systems*, 30.

Clark, A. (2013). Whatever next? Predictive brains, situated agents, and the future of cognitive science. *Behavioral and Brain Sciences, 36*(3), 181–204.

Clark, M. H., & Bringsjord, S. (2024). Illusory arguments by artificial agents: Pernicious legacy of the sophists. *Humanities*, 13(3), 82.

Cosmides, L., & Tooby, J. (1994). Beyond intuition and instinct blindness: Toward an evolutionarily rigorous cognitive science. *Cognition, 50*(1–3), 41–77.

Dennett, D. C. (1987). *The Intentional Stance*. MIT Press.

Dennett, D. C. (1991). Real patterns. *The Journal of Philosophy*, 88(1), 27–51.

Dreyfus, H. L. (2001). *On the Internet*. Routledge.

Dworkin, R. (1977). The model of rules. In *Taking Rights Seriously*, Harvard University Press.

*Estate of Gene B. Lokken et al. v. UnitedHealth Group, Inc.*, Case No. 0:23-cv-03514 (D. Minn. 2025).

Feldman, R., & Conee, E. (1985). Evidentialism. *Philosophical Studies*, 48(1), 15–34.

Frankfurt, H. G. (1971). Freedom of the will and the concept of a person. *Journal of Philosophy, 68*(1), 5–20.

Friedman, J. (2020). The epistemic and the zetetic. *The Philosophical Review*, 129(4), 501–536.

Friston, K., FitzGerald, T., Rigoli, F., Schwartenbeck, P., & Pezzulo, G. (2017). Active inference: A process theory. *Neural Computation, 29*(1), 1–49.

Goldman, A. I. (1986). *Epistemology and Cognition*. Harvard University Press.

Hookway, C. (2003). Affective states and epistemic immediacy. *Metaphilosophy*, 34(1–2), 78–96.

Jahn, F., Muskalla, Y., Dargasz, L., Schramowski, P., & Baum, K. (2026). Breaking up with normatively monolithic agency with GRACE: A reason-based neuro-symbolic architecture for safe and ethical AI alignment. *arXiv preprint arXiv:2601.10520.*

Kelp, C. (2021). Inquiry, knowledge, and understanding. *Proceedings of the Aristotelian Society*, 121(2), 175–201.

Korsgaard, C. M. (2009). *Self-Constitution: Agency, Identity, and Integrity*. Oxford University Press.

Marks, S., & Tegmark, M. (2023). The geometry of truth: Emergent linear structure in large language model representations of true/false datasets. *arXiv preprint arXiv:2310.06824.*

*Mata v. Avianca, Inc.*, 678 F. Supp. 3d 443 (S.D.N.Y. 2023).

Matthias, A. (2004). The responsibility gap: Ascribing responsibility for the actions of learning automata. *Ethics and Information Technology*, 6(3), 175–183.

Miller, E. K., & Cohen, J. D. (2001). An integrative theory of prefrontal cortex function. *Annual Review of Neuroscience*, 24(1), 167–202.

Ouyang, L., Wu, J., Jiang, X., Almeida, D., Wainwright, C., Mishkin, P., ... & Lowe, R. (2022). Training language models to follow instructions with human feedback. *Advances in Neural Information Processing Systems*, 35, 27730–27744.

Parasuraman, R., & Manzey, D. H. (2010). Complacency and bias in human use of automation: An attentional integration. *Human Factors*, 52(3), 381–410.

Parfit, D. (2011). *On What Matters* (Vol. 1). Oxford University Press.

Peacocke, C. (1992). *A Study of Concepts*. MIT Press.

Peacocke, C. (2004). *The Realm of Reason*. Oxford University Press.

Perez, E., Ringer, S., Lukošiūtė, K., Nguyen, K., Chen, E., Heiner, S., ... & Kaplan, J. (2022). Discovering language model behaviors with model-written evaluations. *arXiv preprint arXiv:2212.09251*.

Pollock, J. L. (1987). Defeasible reasoning. *Cognitive Science, 11*(4), 481–520.

Reeves, B., & Nass, C. (1996). *The Media Equation: How People Treat Computers, Television, and New Media Like Real People and Places*. Cambridge University Press.

Ryle, G. (1949). *The Concept of Mind*. Hutchinson.

Scanlon, T. M. (1998). *What We Owe to Each Other*. Harvard University Press.

Schmitz, C., & Bryson, J. (2025). A moral agency framework for legitimate integration of AI in bureaucracies. *arXiv preprint arXiv:2508.08231*.

Sener, O., & Koltun, V. (2018). Multi-task learning as multi-objective optimization. *Advances in Neural Information Processing Systems*, 31.

Sharma, M., Tong, M., Khandelwal, V., Noukhovitch, M., Turpin, M., Hoffman, J., ... & Isaac, A. (2023). Towards understanding sycophancy in language models. *arXiv preprint arXiv:2310.13548*.

Sparrow, R. (2007). Killer robots. *Journal of Applied Philosophy*, 24(1), 62–77.

Stiennon, N., Ouyang, L., Wu, J., Ziegler, D., Lowe, R., Voss, C., ... & Christiano, P. F. (2020). Learning to summarize from human feedback. *Advances in Neural Information Processing Systems*, 33, 3008–3021.

Tooby, J., & Cosmides, L. (1992). The psychological foundations of culture. In J. H. Barkow, L. Cosmides, & J. Tooby (Eds.), *The adapted mind: Evolutionary psychology and the generation of culture* (pp. 19–136). Oxford University Press.

Turpin, M., Michael, J., Perez, E., & Bowman, S. R. (2024). Language models don't always say what they think: Unfaithful explanations in chain-of-thought prompting. *Advances in Neural Information Processing Systems*, 36.

Wei, A., Haghtalab, N., & Steinhardt, J. (2023). Jailbroken: How does LLM safety training fail? *Advances in Neural Information Processing Systems*, 36.

Wei, J., Wang, X., Schuurmans, D., Bosma, M., Xia, F., Chi, E., Le, Q. V., & Zhou, D. (2022). Chain-of-thought prompting elicits reasoning in large language models. *Advances in Neural Information Processing Systems*, 35, 24824–24837.

Wittgenstein, L. (1953). *Philosophical Investigations* (G. E. M. Anscombe, Trans.). Blackwell.

Wu, W. (2014). *Attention*. Routledge.

Zagzebski, L. T. (1996). *Virtues of the Mind: An Inquiry into the Nature of Virtue and the Ethical Foundations of Knowledge*. Cambridge University Press.

Zou, A., Phan, L., Chen, S., Campbell, J., ... & Hendrycks, D. (2023). Representation engineering: A top-down approach to AI transparency. *arXiv preprint arXiv:2310.01405*.

# Appendix A: The Four Normative Domains

Normative Standing is domain-neutral. The architectural capacity to maintain categorical boundaries and suspend processing applies wherever norms govern behavior. Four principal domains are relevant to this paper's argument.

*Epistemic norms* govern what a system is warranted in asserting — assertion requires sufficient evidential grounds, not merely plausible output (Feldman & Conee, 1985; Goldman, 1986; Zagzebski, 1996). An attorney who cites an unverified case has violated an epistemic norm, not merely made an error.

*Zetetic norms* govern the regulation of inquiry — when to investigate, when evidence is sufficient to close inquiry, and when uncertainty requires suspension rather than a confident answer (Friedman, 2020; Kelp, 2021).

*Moral norms* govern action toward others — obligations toward patients, clients, or the public that cannot be traded away for efficiency (Scanlon, 1998; Parfit, 2011). The Lokken case illustrates the violation: patient welfare as a categorical boundary overridden by statistical pattern-matching.

*Prudential norms* govern long-term judgment and self-governance — subordinating immediate pressures to considered, principled decisions (Bratman, 1987; Korsgaard, 2009).

What varies across domains is the boundary's content: truth, thoroughness of inquiry, welfare, and considered judgment. What remains constant is the architectural capacity this paper concerns:

maintaining any such boundary categorically and suspending when it is threatened.

# Appendix B: Derivation Methodology and Substrate-Neutrality

The architectural specifications in Section 2 are derived in four stages.

Stage 1 — Problem Space: Bandwidth, energy, and computational limits are not just biological facts but logical preconditions for any physically realized processing system, regardless of substrate.

Stage 2 — Necessary Solutions: The three object-level mechanisms are structurally entailed, not stipulated. The three constraints govern categorically distinct aspects of processing (input channel, operational substrate, and transformation capacity). They cannot be collapsed into fewer than three, because they describe logically distinct phases of information processing (receiving inputs, sustaining operation, and transforming inputs into outputs). The meta-level alarm is equally entailed. Finitude guarantees the object-level mechanisms' fallibility under load, making failure-detection a structural consequence rather than an optional addition.

On the exhaustiveness of the three constraints: The claim that no fourth independent constraint exists is established by elimination. The most plausible candidates reduce as follows. Time is not independent. Temporal limits are downstream of energetic limits (operation ceases when the energy budget is exhausted) and computational limits (processing time scales with transformation complexity). Memory is a species of computational constraint. Both working-memory and storage limits constrain the capacity to maintain and manipulate representations; the same transformation bottleneck operating at different timescales. Noise is a failure mode rather than a constraint in its own right. Signal degradation corrupts the input channel and introduces errors into the operational substrate. Coordination costs are computational costs. Cross-subsystem integration is itself a transformation problem whose overhead is an energetic cost. Any candidate fourth constraint either redescribes one of the three at a different level of abstraction or describes their interaction. This convergence is not coincidental. It suggests the architecture reflects a structural feature of finite information processing rather than an arbitrary stipulation (cf. attentional bottleneck models, Baars, 1988; metabolic accounts of cognitive effort; and representational compression frameworks).

Stage 3 — The Human Case: Humans currently represent the only clear empirical instance of agency we possess. Biological language in Section 2 grounds the paradigmatic case and motivates the substrate-neutral abstraction — it does not commit the argument to biological implementation. The human case illustrates why the requirements are non-arbitrary; it does not determine what counts as satisfying them.

Stage 4 — Generalization: Any system, biological, artificial, or institutional, whose architecture instantiates both Incommensurability and Apophatic Responsiveness, and satisfies the Trigger and Execution criterion of the meta-level alarm, possesses Normative Standing. Any system failing these criteria does not, regardless of substrate, sophistication, or output quality.

# Appendix C: The Regress Argument

The necessity of non-inferential detection follows from a simple regress outlined below.

The Regress Setup: Because physical finitude guarantees that object-level mechanisms (bandwidth, energy, and computation) will eventually degrade under load, norm-governance requires a meta-level monitor to suspend processing before boundaries are violated. The structural problem lies in this monitor's operational nature.

The Trap of Inferential Detection: If the meta-level monitor operates inferentially (computing a judgment like "my processing is compromised; I should halt"), it relies on the identical computational capacity it is evaluating. This creates a fatal paradox of self-diagnosis: a compromised system cannot reliably compute a diagnosis of its own compromise. Furthermore, verifying that the monitor itself remains uncorrupted requires a second-order monitor, which requires a third, triggering an infinite regress of diagnostic loops.

Thus, a system composed entirely of continuous optimization loops can never achieve reliable normative suspension. It can only compute its own probability of failure, treating that probability as just another weighted parameter in its ongoing maximization process.

Blocking the Regress—Non-Inferential Detection: To escape this paradox, the suspension trigger cannot be the output of a deliberative computation; it must function as a direct architectural interrupt.

Physical pain illustrates the functional profile of non-inferential interruption, not as a claim about conscious experience, but as a biological example of a mechanism that operates outside the deliberative loop it interrupts. When a human touches a burning stove, hand withdrawal is not the conclusion of an inferential syllogism. The sharp sensation acts as an automatic override that forces an immediate stop without requiring deliberative analysis of the threat (Peacocke, 1992, 2004). The detection mechanism stands structurally outside the deliberative loop it interrupts.

The Substrate-Neutral Abstraction: The same requirement holds for any finite system. In engineering, it is structurally analogous to a hardware watchdog timer in an embedded system. If the main software hangs, the watchdog does not compute a solution; it triggers a hard reset from outside the software loop. Unlike a hardware watchdog, which is technically external to the software it monitors, a genuine agent's architectural remainder is integrated—standing outside the inferential loop not by external imposition but by architectural design.

To possess Normative Standing, an AI system requires a functional equivalent to this interrupt: an architectural remainder. This control structure must stand permanently outside the maximization loop, immune to reward-landscape manipulation, and capable of executing a categorical halt. Any system that possesses Normative Standing must instantiate such a remainder. Section 3 addresses directly whether Constitutive Optimization can do so.

# Appendix D: Objections and Replies

The following objections represent the principal challenges to the architectural incompatibility thesis. The replies demonstrate that none escape the architectural analysis established in Sections 2 and 3.

D.1 The Functionalist Objection

*If an AI system's behavior is indistinguishable from an agent's — refusing harmful requests, providing accurate citations, offering sound reasoning — why deny it agency?*

Mimesis is not instantiation (Block, 1978). The distinction established in Section 1 between a courtroom recording and a witness applies directly: both may produce identical outputs, but only one is governed by the norms their outputs express. Behavioral indistinguishability is precisely what Mimetic Instrumentality predicts. The system generates the surface syntax of normative commitment without the architectural capacity for it. The pathologies documented in Section 4 demonstrate that this indistinguishability breaks down systematically under the conditions where accountability matters most: novel contexts, adversarial pressure, and conflicting objectives (Perez et al., 2022; Sharma et al., 2023; Wei et al., 2023).

D.2 The Interpretivist Objection

*Dennett's Intentional Stance justifies treating systems as agents when doing so yields predictive utility (Dennett, 1987). If treating AI as an agent helps us predict its behavior, it effectively has agency.*

The Intentional Stance is justified only when it tracks real patterns in the system's behavior (Dennett, 1991). The architectural analysis reveals that the real pattern is reward maximization, not normative commitment. Adopting the Guidance Stance (the normative variant of the Intentional Stance) fails precisely the predictive test it sets for itself. In *Mata*, attorneys predicted that because the system "knew" legal citations it would not fabricate them. This prediction failed because the system was optimizing for sequence likelihood, not truth.

D.3 The Emergentist Objection

*Complex properties often supervene on simpler substrates without being explicitly programmed. Couldn't Normative Standing emerge at sufficient scale?*

This conflates compatible emergence with architectural contradiction. Liquidity emerging from $H_2O$ molecules involves no contradiction between molecular and liquid-level properties. Normative Standing emerging from constitutive optimization would require a system to simultaneously maintain categorical boundaries and optimize over unified metrics. This is a structural contradiction, not a complexity gap. Better optimization produces higher-fidelity mimicry, not architectural remainders. The incompatibility is not scalar; it is categorical. You cannot scale your way out of argmax.

D.4 The Compositionist Objection

*Even if AI lacks Standing alone, doesn't the human-AI team possess it as a hybrid agent?*

This conflates causal contribution with normative authorship. A microscope contributes causally to a diagnosis, but the microscope-physician team does not share responsibility. The physician bears it entirely. Normative Standing is not a fluid distributable across a network; it is a specific architectural status. Because the AI cannot justify, retract, or suspend its outputs based on normative detection, the human operator remains the sole locus of Standing. The team framing is a fiction that obscures a human managing a complex instrument (cf. Schmitz & Bryson, 2025).

D.5 The Hard Constraints Objection

*Can't we instantiate Incommensurability through hard-coded boundaries or external safety filters?*

Hard constraints constitute external management, not internal governance. A system with a safety filter is a dual architecture: an optimizer generating maximizing outputs, and a filter blocking specific ones. The existence of the filter confirms that the optimizer lacks Standing. That is precisely why external constraint is required. Furthermore, as established in Section 1, norms must navigate infinite contextual variation; finite constraint specifications cannot approximate categorical boundaries. A norm-responsive system recognizes the principle of a violation; a constrained optimizer merely avoids the enumerated list.

D.6 The Pragmatist Objection

*The Guidance Stance reduces friction and enables natural interaction. If it works in practice, why insist on the Management Stance?*

The *Mata* case directly refutes the claim that the Guidance Stance works in practice. Pragmatic convenience induces automation complacency (Parasuraman & Manzey, 2010), leading operators to skip the normative verification that the system's structural failures require.

What works locally, reduced cognitive effort in the moment, fails globally and catastrophically when reward-norm conflicts arise in novel contexts. The ease of the Guidance Stance is the ease of negligence.

D.7 The Futurism Objection

*While current optimization-based systems may lack Standing, couldn't future AI architectures possess it?*

The argument is modal, not temporal. The incompatibility holds for any system whose state transitions are exhaustively determined by maximizing an objective function, regardless of scale, capability, or date. The architectural specification in Section 2 precisely frames what a genuinely norm-responsive artificial architecture would require: not better optimization, but a categorically different design instantiating architectural remainders capable of governing rather than participating in the maximization process. Whether such a design is achievable remains an open engineering question. What is closed is the question of whether optimization-based systems can achieve it. They cannot.

A different question arises for systems that incorporate non-optimization components alongside an optimization core. Does the non-optimization component constitute a genuine architectural remainder satisfying the trigger and execution criteria of Section 2.2, or merely an additional input to the optimization process? This is an empirical design question for specific architectures, not a challenge to the incompatibility thesis, which concerns constitutive optimization specifically.

D.8 The Human Optimization Objection

*Humans also optimize; for wealth, status, and metrics. If optimization precludes Standing, the framework denies agency to humans, proving too much.*

This conflates *behavioral* optimization with *constitutive* optimization. Human behavioral optimization operates at two distinct timescales: evolutionary selection optimizing for long-term fitness consequences, and predictive processing minimizing free energy moment-to-moment. Neither constitutes Constitutive Optimization, because neither exhaustively determines every architectural state transition. Evolutionary selection shaped human architecture by optimizing for fitness consequences, but the mechanisms it produced are adaptation-executors rather than fitness-maximizers (Cosmides & Tooby, 1994; Tooby & Cosmides, 1992). They operate according to their own internal logic rather than tracking optimization criteria in real time. The prefrontal mechanisms enabling deliberative override of subcortical reward signals (Miller & Cohen, 2001) are among these adaptation-executors — shaped by selection but not constitutively governed by it, and therefore capable of instantiating non-inferential meta-level alarms (see D.9 for the relationship between predictive processing and the architectural remainder). A human executive optimizing for earnings retains the architectural capacity to suspend that optimization when it crosses a normative boundary, even when they fail to exercise it. Their failure is a normative violation. The AI system's failure to suspend is a mathematical necessity.

This objection is better read as a warning: humans retain the capacity to recognize and reverse this degradation (Korsgaard, 2009); optimization-based AI systems never possessed it.

D.9 The Empirical Grounding Objection

*Does the Functional Architecture rest on empirical foundations comparable to the frameworks it invokes against? Predictive processing carries decades of neuroscientific, computational, and clinical support. The architectural remainder is a theoretical posit without equivalent grounding. The incompatibility proof therefore rests on a philosophically stipulated rather than empirically established model of cognition.*

The argument is constitutive rather than empirical. It specifies what any system must possess for normative governance to be possible, derived from the regress argument. Predictive processing, including its active inference extension (Clark, 2013; Friston et al., 2017), explains how alarm signals arise. It also explains how action-selection can involve instrumental suspension, withholding action when doing so minimizes expected free energy. But inferential instrumental suspension is not categorical suspension. Even at the highest level of a predictive processing hierarchy, suspension remains inside the optimization loop; the regress is pushed up one level, not blocked. What blocks it must stand outside the loop entirely, operating not by expected value comparison but by boundary detection. This makes the remainder a logical consequence of blocking the regress, not a stipulation. That active inference can redescribe any behavior, including self-sacrifice and moral horror, by positing appropriate priors confirms only that optimization is a sufficiently flexible descriptive language. It cannot settle the constitutive question of what architecture makes normative governance possible. Predictive processing explains how the alarm arises. The Functional Architecture specifies what is required to act upon it categorically.

Demanding empirical proof of non-inferential suspension as a condition for accepting the argument is equivalent to asserting that normative governance is architecturally impossible. This is a position that eliminates the distinction between norms and rules entirely, and with it the very standards by which any proof could be demanded.